\documentclass[sigconf]{acmart}  %

\definecolor{blue}{rgb}{0.21,0.49,0.74}
\hypersetup{breaklinks,colorlinks,allcolors=blue}

\usepackage{times}
\usepackage[T1]{fontenc}
\usepackage[utf8]{inputenc}
\usepackage{microtype}
\usepackage{graphicx}
\usepackage{booktabs}
\usepackage{xcolor}

\usepackage{multirow}
\usepackage{colortbl}
\usepackage[capitalize]{cleveref}

\title{Memory-enhanced Retrieval Augmentation for Long Video Understanding}

\begin{document}

\author{\textbf{Huaying Yuan\textsuperscript{1}}, \textbf{Zheng Liu\textsuperscript{2}}, \textbf{Minhao Qin\textsuperscript{4}}, \textbf{Hongjin Qian\textsuperscript{2}}, \textbf{Y Shu\textsuperscript{3}}, \\
 \textbf{Zhicheng Dou\textsuperscript{1*}}, \textbf{Ji-Rong Wen\textsuperscript{1*}}, \textbf{Nicu Sebe\textsuperscript{3}}  \\
 \textsuperscript{1}Gaoling School of Artificial Intelligence, Renmin University of China  \\
 \textsuperscript{2}Beijing Academy of Artificial Intelligence, 
  \textsuperscript{3}University of Trento, \\
  \textsuperscript{4}Institute of Automation, Chinese Academy of Sciences\\
}

\begin{abstract}

Efficient long-video understanding~(LVU) remains a challenging task in computer vision. Current long-context vision-language models~(LVLMs) suffer from information loss due to compression and brute-force downsampling. While retrieval-augmented generation (RAG) methods mitigate this issue, their applicability is limited due to explicit query dependency. To overcome this challenge, we introduce a novel memory-enhanced RAG-based approach called \textbf{MemVid}, which is inspired by the cognitive memory of human beings. Our approach operates in four basic steps: 1) \textbf{memorizing} holistic video information, 2) \textbf{reasoning} about the task's information needs based on memory, 3) \textbf{retrieving} critical moments based on the information needs, and 4) \textbf{focusing} on the retrieved moments to produce the final answer. To enhance the system's memory-grounded reasoning capabilities while achieving optimal end-to-end performance, we propose a \textbf{curriculum learning} strategy. This approach begins with supervised learning on well-annotated reasoning results, then progressively explores and reinforces more plausible reasoning outcomes through reinforcement learning. We perform extensive evaluations on popular LVU benchmarks, including MLVU, VideoMME and LVBench. In our experiments, MemVid demonstrates superior efficiency and effectiveness compared to both LVLMs and RAG methods. 
\end{abstract}

\maketitle

\begin{figure*}[]
    \centering
    \includegraphics[width=\textwidth]{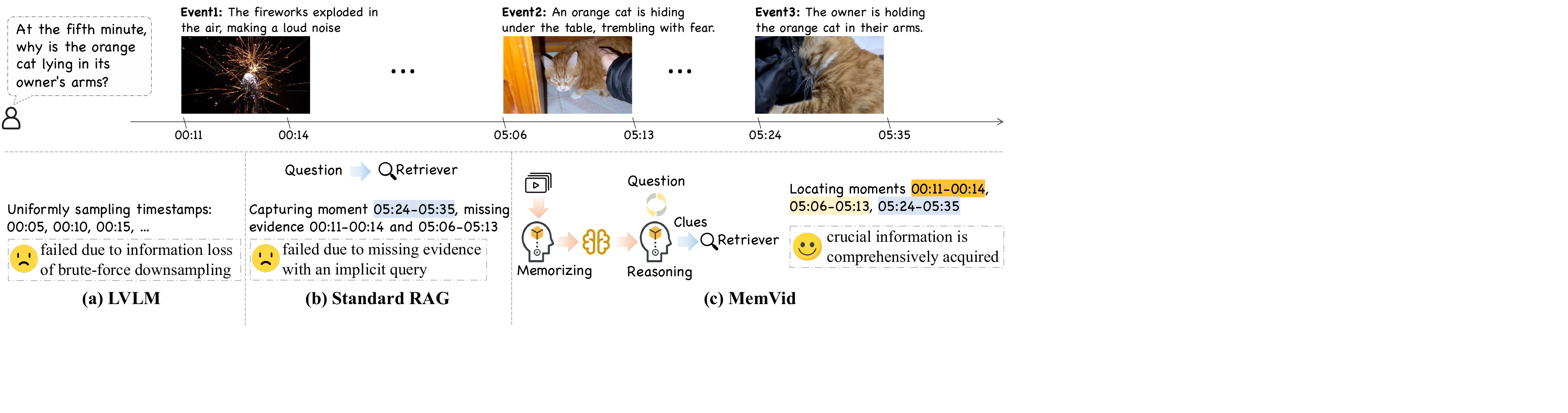}
    \vspace{-0.6cm}
    \caption{Comparison of different frameworks. MemVid outperforms current methods by generating useful clues based on holistic memory, facilitating the location of comprehensive details for question answering.}
    \label{fig:compare}
        \vspace{-0.4cm}
\end{figure*}

\section{Introduction}\label{sec:intro} 

Long-video understanding plays a crucial role in real-world applications such as video analysis, autonomous driving, and embodied AI. However, this task presents significant challenges for traditional large vision-language models (VLMs)~\cite{internvl,qwen2vl,bai2023qwenvl}, which are primarily designed for short visual inputs, including single images, multiple images, or short videos. Recent work \cite{longva,longvila,shu2024videoxl} has expanded the context window of VLMs, leading to LVLMs. Despite this progress, these methods still remain limited by information loss from brute-force downsampling and compression, as well as high computational costs when processing long video sequences.

Retrieval-augmented generation (RAG) has emerged as a promising approach for handling long-sequence problems. By retrieving useful information from long-sequence data, the model can perform its generation based on a highly simplified input, thus enabling cost-effective fulfillment of its task. While standard RAG methods excel at addressing clearly specified queries, such as ``\textit{When does the orange cat lie in its owner's arms?}'', they are insufficient for general long-video understanding problems, which often involve implicit and complex information needs. As shown in Figure~\ref{fig:compare}(b), consider the query ``\textit{At the fifth minute, why is the orange cat lying in its owner's arms?}'' Instead of directly resorting to a retrieval model with Event3 (as mentioned in the question), the model must identify the cat's preceding behavior (Event2) and infer the implicit relationship between the fireworks explosion (Event1) and the cat's reaction described in the question. This process requires reasoning beyond straightforward retrieval, highlighting the limitations of conventional RAG techniques in handling dispersed and context-dependent information in long videos.  

In contrast, humans tackle long-video understanding problems far more effectively. They first go through the entire video, forming the memorization of the overall content. When faced with a specific question, they reason about the problem to determine what information is required. Only then do they retrieve key moments from memory, focusing on those relevant details to arrive at a final answer. This structured process, including integrating comprehension, reasoning, and targeted retrieval, enables humans to handle complex long-video understanding tasks with remarkable proficiency.

With the above inspiration, we propose a novel RAG framework for long-video understanding, called \textbf{MemVid} (\textbf{Mem}ory-enhanced retrieval augmentation for long \textbf{Vid}eo understanding). As shown in Figure~\ref{fig:compare}(c), MemVid operates in four basic steps. First, it generates the memory for the holistic information of the long video. Second, it reasons about the information needs for a concrete problem based on its memory. Third, it retrieves crucial moments from the video as required by the information needs. And lastly, it generates the final answer based on the retrieval results. 

The above workflow is driven by three essential modules: the memorizer, the retriever, and the generator. In our work, we focus on optimizing the memorizer while keeping other modules fixed. To achieve optimal end-to-end performance, we introduce a curriculum learning framework. Our training process begins with supervised learning, where the memorizer is trained to generate well-structured reasoning outputs based on high-quality annotations obtained from powerful long-video VLMs. Once this foundation is completed, the memorizer explores various candidate reasoning trajectories, reinforcing those that lead to high-quality answers. This approach ensures a progressive refinement of reasoning capabilities, ultimately enhancing the system's overall performance.

To assess the effectiveness of MemVid, we conduct comprehensive experiments using various long-video understanding (LVU) benchmarks, including VideoMME\cite{videomme}, MLVU\cite{zhou2024mlvu}, and LVBench~\cite{wang2024lvbench}. In our experiment, MemVid achieves notable advantages against existing RAG-based LVU methods, while significantly improves the cost-effectiveness in comparison to popular long-video VLMs. 

In summary, our contributions are threefold:
\begin{enumerate}
    \item We introduce MemVid, a novel retrieval augmentation framework tailored for long-video understanding. To the best of our knowledge, this is the first approach of its kind, emphasizing the critical role of reasoning and memorization in comprehending long videos. 
    \item We design an effective curriculum learning framework that enhances the memorizer's ability to improve end-to-end RAG performance by leveraging diverse training signals.  
    \item We conduct extensive experiments, which showcase MemVid's ability to achieve high-quality results while significantly improving the cost-effectiveness in long-video understanding.  
\end{enumerate}

\section{Related Work}
\subsection{Large Vision-language Models}

The remarkable advancements in large language models (LLMs) have catalyzed a growing interest in the development of comprehensive multi-modal artificial general intelligence systems, with numerous pioneering studies~\cite{internvl,QwenLM,liu2023visual_llava,zhu2023minigpt4} making significant contributions to this emerging field. Building upon foundational work, researchers have explored various architectural approaches and training methodologies to bridge the gap between visual and linguistic understanding. InternVL~\cite{internvl} stands out as a particularly influential model, pushing the boundaries by scaling vision models to an impressive 6 billion parameters and achieving robust alignment with LLMs through extensive multimodal pretraining. Parallel developments have emerged through frameworks like QwenVL~\cite{bai2023qwenvl}, which extends the capabilities of the QwenLM~\cite{QwenLM} architecture by incorporating several key innovations including a purpose-built visual encoder, a unified interface for both input and output processing, a sophisticated three-phase training regimen, and a comprehensive multilingual dataset specifically designed for multimodal tasks.

As the field progresses, research focus has naturally expanded beyond static image analysis to encompass the more complex domain of video understanding. The predominant approach in current systems involves a two-stage process where video frames are first encoded into visual representations, which are then transformed into tokenized formats suitable for processing by large language models. ST-LLM~\cite{stllm} exemplifies this paradigm while introducing the novel approach of directly feeding raw spatiotemporal tokens to LLMs, demonstrating surprisingly effective modeling of video sequences. Alternative methodologies have emerged through projects like VideoChat~\cite{li2023videochat}, which achieves notable improvements in temporal reasoning and event comprehension through specialized trainable neural adapters, with its successor VideoChat2~\cite{li2024mvbench} further enhancing performance via comprehensive instruction tuning across diverse scenarios. Early implementations predominantly relied on BLIP-2's Q-Former~\cite{li2023blip2} architecture for feature fusion, though recent trends have favored more streamlined designs. This shift is evident in models such as VideoLLaVA~\cite{videollava} and MiniGPT4-Video~\cite{minigpt4video-2024}, which implement straightforward linear transformations to project visual features into the linguistic space. Other innovative approaches, including those employed by Video-ChatGPT~\cite{maaz2023videochatgpt} and Valley~\cite{luo2023valley}, utilize sophisticated token pooling mechanisms to extract and concentrate the most salient visual information. The critical importance of data quality has become increasingly apparent in recent studies. LLaVA-Video~\cite{zhang2024llavavideo} highlights how carefully curated synthetic training data can dramatically improve a model's ability to follow video instructions accurately. Similarly, Qwen2VL~\cite{qwen2vl} demonstrates remarkable versatility across diverse tasks, achieved through two primary innovations: massive scaling of training data and the implementation of adaptive resolution processing techniques. 

Despite these advances, current multi-modal systems face inherent limitations in processing long videos, primarily constrained by context length restrictions (typically capped at 128 frames). This remains a fundamental challenge between the research community and practical applications.

\subsection{Long Large Vision-language Models}

Unlike short videos, long videos often contain significant temporal redundancy for a specific task, with only certain segments being relevant for generating meaningful content. Existing approaches typically address this redundancy through memory mechanisms or compression modules. MovieChat~\cite{moviechat2023} and MA-LMM~\cite{malmm2024} use memory banks to store historical visual features and compress them using a consolidation strategy. To enhance the model’s capacity for understanding long-range context, LLAMA-VID~\cite{llama-vid2023} compresses each frame into just two tokens with a context attention module. LongVLM~\cite{weng2024longvlm} introduces a token-merging module that combines local features from short segments with global context features, while Video-CCAM~\cite{fei2024videoccam} utilizes cross-attention mechanisms with causal cross-attention masks. Some approaches~\cite{longva, longvila, wang2024longllava} directly enhance the model's inherent capability to comprehend long video inputs. While these models extend the processing length for long videos, they make a suboptimal trade-off, struggling with redundancy and quadratic computational overhead.

\subsection{Retrieval-augmented Video Understanding}
Retrieval Augmented Generation (RAG) is a classic enhancement technique in the text domain that uses a retriever to obtain relevant context for content generation~\cite{gao2023retrieval,huang2024survey, zhao2024retrieval, fan2024survey,LLM4IRSurvey}. Its advantage lies in the ability to extract key information from vast amounts of irrelevant data, effectively enabling long-range modeling. Due to the temporal redundancy in long videos, RAG is a natural fit for such scenarios. Some studies have explored the feasibility of applying RAG for long video understanding. DrVideo~\cite{ma2024drvideo} and Goldfish~\cite{ataallah2024goldfish} generate text captions for key frames and video clips, respectively, and implement a RAG framework based on these captions. However, this approach creates a significant gap between raw video content and textual captions, which hampers effective retrieval and limits video understanding. Furthermore, due to the lack of contextual and deep semantic information, directly using queries to retrieve relevant content from the video is suboptimal. To address this limitation, we propose a generative context-aware query expansion mechanism that operates directly on the video content. This mechanism improves retrieval performance by expanding the queries to include more contextual information, enabling accurate responses within a limited input length, focusing on key content, while also reducing computational costs by limiting the amount of data processed.

\begin{figure}[]
    \raggedright
\includegraphics[width=0.9\linewidth]{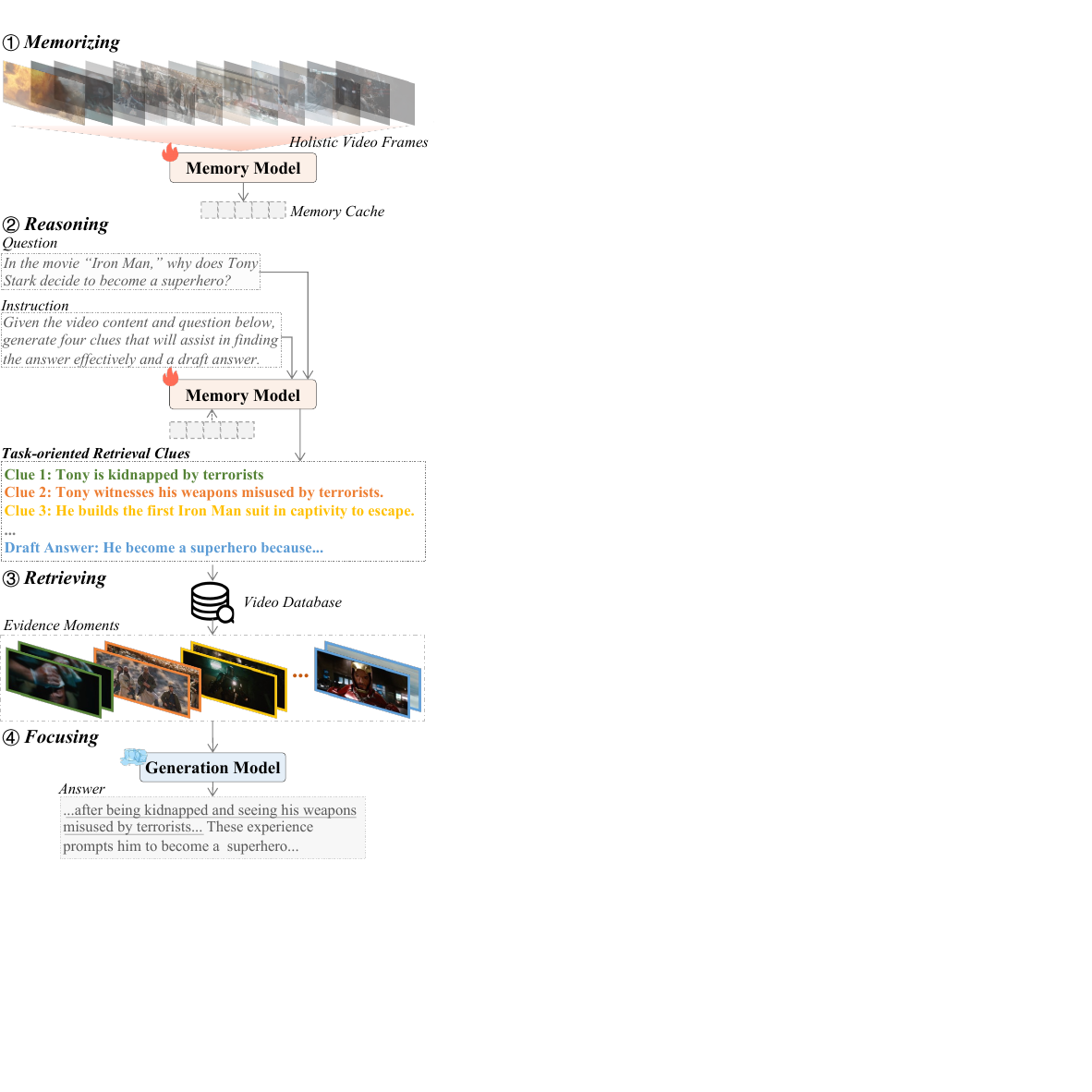}
    \vspace{-0.3cm}
        \caption{Illustration of MemVid's four-step process. Evidentiary moments are color-coded to align with their corresponding clues. MemVid generates outputs from a highly simplified input, ensuring cost-effective task execution.}
    \label{fig:main}
    \vspace{-0.4cm}
\end{figure}

\section{Methodology}

\subsection{Overview of MemVid}
\label{sec:overview}
Long video understanding aims at answering a question \( Q \) about a video \(\mathcal{V}\) composed of \(N\) frames. Current VLMs naively downsample a sparse subset \(\mathcal{S}\) of \(k\) frames  (e.g., uniform/dense sampling) from \(\mathcal{V}\) for answer generation:  
\begin{equation}
A = \mathcal{G}(\mathcal{S}, Q \,|\, \theta),
\end{equation}
where \(\mathcal{G}\) denotes the generation model with parameters \(\theta\). Such brute-force downsampling approach leads to significant information loss for long videos due to \(k \ll N\). To mitigate this, LVLMs attempt to increase \(k\) (e.g., to 1,024 frames) to reduce information loss. However, for hour-long videos (which typically contain around 72K natural frames), this remains far from sufficient. To address this limitation, RAG methods adopt a retrieval strategy to identify and extract task-relevant moments:   
\begin{equation}
\mathcal{S}' = \text{Top-}k\left( Q \,|\, \mathscr{C}, \omega\right),
\end{equation}
where \(\omega\) denote the retriever's parameters, and \(\mathscr{C}\) represent the retrieval database, which consists of moments or frames extracted from the original video \(\mathcal{V}\). The \(\text{Top-}k(\cdot)\) function retrieves the \(k\) moments most relevant to the query \(Q\) from \(\mathcal{V}\). Downstream VLMs generates answers based on these task-relevant key frames:
\begin{equation}
A' = \mathcal{G}(\mathcal{S}', Q \,|\, \theta).
\end{equation}
Despite advancements, standard RAG systems exhibit limitations in handling implicit or complex queries due to insufficient reasoning capabilities, which often leads to suboptimal retrieval and missed evidence, degrading answer quality. To tackle these challenges, we propose \textbf{MemVid}, a memory-enhanced framework that emulates human-like cognition by constructing a global video memory to guide context-aware retrieval. As shown in Figure~\ref{fig:main}, MemVid operates in four stages:  

\textbf{1. Memorizing:} Scan the video and store it into memory \(\mathcal{M}\) using a memory model to maintain a holistic understanding:
\begin{equation}
\mathcal{M} = \mathcal{R}(\mathcal{V'} \,|\, \phi),
\end{equation}
where \(\mathcal{R}\) is the memory model parameterized by \(\phi\) and \(\mathcal{V'}\) is uniformly down-sampled frames of the original video.  

\textbf{2. Reasoning:} Given a specific question \(Q\), leverage the memory model to infer task-oriented retrieval clues \(\mathcal{C}\) by reasoning over \(Q\) and the holistic memory \(\mathcal{M}\):  
\begin{equation}
\mathcal{C} = \mathcal{R}(Q \,|\, \mathcal{M}, \phi).
\end{equation}
Here, the task-oriented retrieval clues consist of a set of clues along with a draft answer for the question: \(\mathcal{C} = \{c_1, \ldots, c_m, c_a\}\).

\textbf{3. Retrieving:} Segment long video into moments  \(\mathscr{C} = \{s_1,\ldots,s_M\}\) and retrieve moments relevant to each retrieval clue \(c \in \mathcal{C}\) and aggregate retrieval results:  
\begin{equation}\label{equation:retriever}
\mathcal{S}'' = \bigcup_{c \in \mathcal{C}} \text{Top-}k\left(c \,|\mathscr{C},\, \omega\right).
\end{equation}

\textbf{4. Focusing:} Generate the final answer based on retrieved informative evidentiary moments \(\mathcal{S}''\) and the original question:  
\begin{equation}
A'' = \mathcal{G}(\mathcal{S}'', Q \,|\, \theta).
\end{equation}
Here, the retrieved moments are temporally reordered and uniformly sampled to meet the downstream context constraints.

By incorporating a memory module to perform both memorizing and reasoning tasks, MemVid overcomes the information loss of brute-force downsampling and evidence missing of standard retrieval, improving effectiveness in long video understanding.

\subsection{Reasoning-oriented Memory Module}  
\label{sec:memory}  
The memory module is expected to have three characteristics: (1) supporting a holistic view of the video, (2) facilitating the reasoning process, and (3) maintaining flexibility for generating retrieval clues for downstream tasks. To achieve this, we design a flexible memory module based on key-value (KV) cache. Specifically, given an input video \(\mathcal{V'} \in \mathbb{R}^{T \times H \times W \times 3}\), where \(T\) frames are uniformly sampled from the original video, a pretrained visual encoder \(E_v\) is used to compress raw video content into token-like visual features:
\begin{equation}
    F = E_v(\mathcal{V'})  \in \mathbb{R}^{(T \times K) \times d_v},
\end{equation}
where \(K\) denotes the token number of each frame, and \(d_v\) the feature depth. While \(F\) captures holistic spatial-temporal information, it lacks reasoning capabilities.

\begin{sloppypar}
To overcome this, we further process visual features \(F\) into reasoning-oriented KV caches using a causal transformer-based language model \(\Theta\). We convert reasoning instructions into token embeddings using the embedder \(E_q\), obtaining \(\{x_1, \ldots, x_p\}\), while the token-like visual features are denoted as \(\{x_{p+1}, \ldots, x_{p+T\times K+1}\}\), and the total input can be represented as \(X = \{x_1,...,x_{p+T\times K+1}\}\). The input \(X\) is processed by a casual transformer. For each timestep \(t \in [1, p+T\times K+1]\), we compute the key and value as:
\end{sloppypar}
\begin{equation}
K_t = \mathcal{W}_k X_t, \quad V_t = \mathcal{W}_v X_t.
\end{equation}
The KV cache is then updated by concatenating the new key-value pairs with the previous ones:
\begin{equation}
K \leftarrow \text{Concat}(K,K_t), \quad V \leftarrow \text{Concat}(V,V_t).
\end{equation}
The resulting memory comprises all KV states from both reasoning instructions and visual features:  $\mathcal{M} = \{K, V\}.$ 

When a question \(Q\) arrives, we dynamically concatenate the memory \(\mathcal{M}\) with the question embedding \(E_q(Q)\) to reason out useful clues for evidence retrieval:  
\begin{equation}
\mathcal{C} = \mathcal{R}\left(\text{Concat}(\mathcal{M}; E_q(Q))\right),
\end{equation}
As illustrated in Figure~\ref{fig:main}, this design addresses the narrow applicability of standard RAG methods by enabling dynamic localization of evidentiary moments for \(Q\) through holistic reasoning over the video’s memory.

\begin{figure}[]
    \centering
    \includegraphics[width=0.8\linewidth]{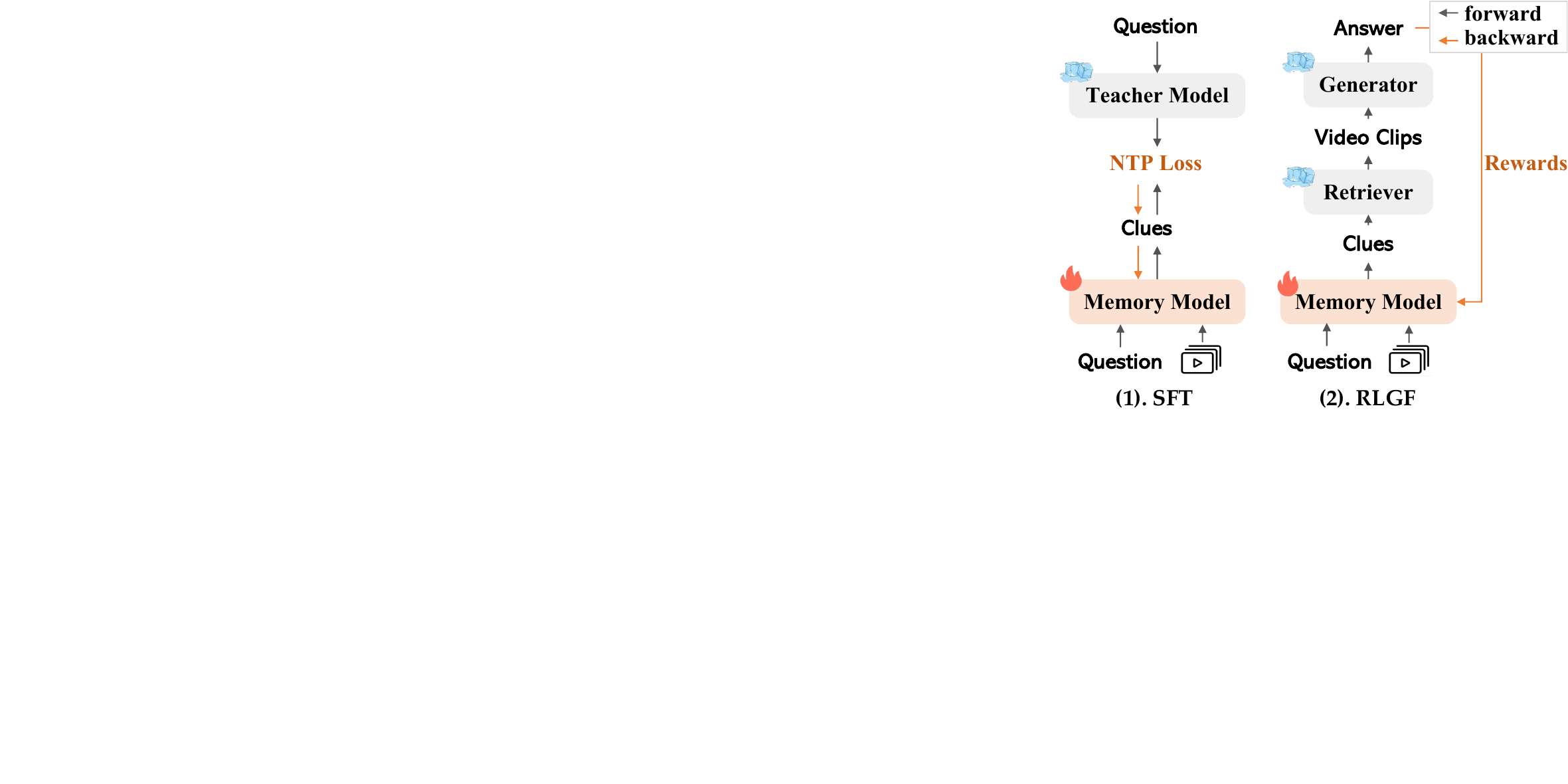}
    \vspace{-0.2cm}
    \caption{Curriculum learning framework combining (1) supervised finetuning warmup and (2) reinforcement learning with generation feedback for end-to-end optimization.}
    \label{fig:pipe}
    \vspace{-0.3cm}
\end{figure}

\subsection{Clue-guided Retrieval}  
To retrieve relevant video segments, we first construct a searchable candidate pool by splitting the long video into non-overlapping, fixed-duration moments. This approach avoids modality loss from dense captioning~\cite{ataallah2024goldfish,ma2024drvideo} and temporal disruption from frame-based splitting. For retrieval, we compute textual embeddings for each clue \(c \in \mathcal{C}\) and moment \(s_j \in \mathscr{C}\) using a pretrained video retriever~\cite{zhu2023languagebind}. The moments are ranked by cosine similarity, and the top-\(k\) are aggregated in temporal order. To balance local and global context, we sample \(\alpha\) frames from retrieved moments and \(1-\alpha\) uniformly from the full video, then feed them into the frozen generation model.

\subsection{Curriculum Learning Framework}
Within the memorizer-retriever-generator framework, our primary focus is on optimizing the memorizer while freezing the other two components. The memorizer generates intermediate retrieval clues without ground truth labels, which poses a challenge for training. To tackle this problem, we propose a curriculum learning framework. In the first stage, we leverage a powerful teacher model to generate well-structured retrieval clues to warm up the memorizer, helping it quickly understand the task. Building upon this, we further propose reinforcement learning with generation feedback, which uses the correctness of the downstream generator's answers to align retrieval clues with optimal end-to-end performance.

\paragraph{Supervised Fine-Tuning Warmup.}  
Our training process begins with supervised learning, where the memorizer is trained to generate well-structured reasoning outputs based on high-quality annotations obtained from powerful long-video VLMs. For synthetic data generation, we employ a powerful 72B-parameter VLM to produce structured reasoning outputs. We then filter these outputs, retaining only those that yield correct downstream answers, ensuring high-quality training data. Finally, the memorizer is optimized via next-token prediction on this refined dataset:
\begin{equation}
\mathcal{L}_{\text{NTP}}(\theta) = -\sum_{t=1}^{T} \log P_\theta(w_t \mid w_{<t}),
\end{equation}  
where \( P_\theta(w_t \mid w_{<t}) \) predicts token \( w_t \) given context \( w_{<t} \). The resulting model demonstrates robust task comprehension and structured clue generation.

\paragraph{Reinforcement Learning with Generation Feedback.}  
To achieve optimal end-to-end performance, we refine clue quality through direct preference optimization (DPO)~\cite{rafailov2024dpo} after the SFT initialization. For each query, we sample multiple clues from the model and rank them by the generator's correctness probability. Preference pairs \((y_i^+, y_i^-)\) are constructed with a minimum score margin \(\tau\) to ensure quality distinction. The optimization follows:  
\begin{equation}
\mathcal{L}_\text{dpo} = -\sum_{i} \log \sigma\left( \beta \cdot \left[ \log \frac{\pi_\theta(y_i^+)}{\pi_\text{S}(y_i^+)} - \log \frac{\pi_\theta(y_i^-)}{\pi_\text{S}(y_i^-)} \right] \right),
\end{equation}  
where \(\pi_\theta\) is the learnable policy, \(\pi_\text{S}\) the frozen SFT reference model, and \(\beta\) controls policy divergence. This dual-stage approach accelerates convergence while refining clues with downstream objectives, achieving stronger generalization than single-stage methods.

\begin{table*}[t!]
\centering
\addtolength\tabcolsep{-2.4pt} 
\small
\caption{Experimental results on MLVU-test and VideoMME. $\dag$ indicates that results are reproduced using their official weights.} 
\label{tab:main} 
\vspace{-0.2cm}
\begin{tabular}{lc|c|c|cccc|cccc|c}
\toprule
\multicolumn{1}{c}{\multirow{2}{*}{\textbf{Model}}} & \multicolumn{1}{c}{\multirow{2}{*}{\textbf{Size}}} & \multicolumn{1}{c}{\multirow{2}{*}{\textbf{\# Frames}}} & \multicolumn{1}{c|}{\textbf{MLVU}} & \multicolumn{4}{c|}{\textbf{VideoMME w/o subtitle}} & \multicolumn{4}{c|}{\textbf{VideoMME w subtitle}} &\multicolumn{1}{c}{\multirow{2}{*}{\textbf{Avg}}} \\
\multicolumn{1}{c}{} & \multicolumn{1}{c}{} & \multicolumn{1}{c}{} & \textbf{M-avg} & \textbf{Short} & \textbf{Medium} & \textbf{Long} & \textbf{Avg} & \textbf{Short} & \textbf{Medium} & \textbf{Long} & \textbf{Avg} \\ \midrule
\rowcolor{gray!15}\multicolumn{13}{c}{\textcolor{gray}{\textit{Proprietary Models}}} \\
\textcolor{lightgray}{GPT-4V~\cite{openai2023gpt4}} & \textcolor{lightgray}{-} & \textcolor{lightgray}{10} & \textcolor{lightgray}{43.3} & \textcolor{lightgray}{70.5} & \textcolor{lightgray}{55.8} & \textcolor{lightgray}{53.5} & \textcolor{lightgray}{59.9} & \textcolor{lightgray}{73.2} & \textcolor{lightgray}{59.7} & \textcolor{lightgray}{56.9} & \textcolor{lightgray}{63.3} & \textcolor{lightgray}{55.5} \\
\textcolor{lightgray}{GPT-4o(20240615)~\cite{gpt4o}} & \textcolor{lightgray}{-} & \textcolor{lightgray}{384} & \textcolor{lightgray}{54.9} & \textcolor{lightgray}{80.0} & \textcolor{lightgray}{70.3} & \textcolor{lightgray}{65.3} & \textcolor{lightgray}{71.9} & \textcolor{lightgray}{82.8} & \textcolor{lightgray}{76.6} & \textcolor{lightgray}{72.1} & \textcolor{lightgray}{77.2} & \textcolor{lightgray}{68.0}\\
\textcolor{lightgray}{Gemini-1.5-Pro~\cite{reid2024gemini}} & \textcolor{lightgray}{-} & \textcolor{lightgray}{1/0.5 fps} & \textcolor{lightgray}{-} & \textcolor{lightgray}{{81.7}} & \textcolor{lightgray}{{74.3}} & \textcolor{lightgray}{{67.4}} & \textcolor{lightgray}{{75.0}} & \textcolor{lightgray}{{84.5}} & \textcolor{lightgray}{{81.0}} & \textcolor{lightgray}{{77.4}} & \textcolor{lightgray}{{81.3}} & \textcolor{lightgray}{-}\\ 
\rowcolor{gray!15}\multicolumn{13}{c}{\textit{Open-source VLMs}} \\
VideoChat2~\cite{li2023videochat} & 7B & 16 & 35.1 & 48.3 & 37.0 & 33.2 & 39.5 & 52.8 & 39.4 & 39.2 & 43.8 & 39.5 \\
ShareGPT4Video~\cite{chen2024sharegpt4video} & 7B & 16 & 33.8 & 48.3 & 36.3 & 35.0 & 39.9 & 53.6 & 39.3 & 37.9 & 43.6 & 39.1 \\
Qwen2VL$\dag$~\cite{qwen2vl} & 7B & 128 & 49.6 & 68.0 & 58.4 & 47.9 & 58.1 & 70.7 & 66.2 & 53.4 & 63.4 & 57.0 \\
InternVL-Chat-V1.5~\cite{internvl} & 20B & 10 & 37.3 & 60.2 & 46.4 & 45.6 & 47.8 & 61.7 & 49.1 & 46.6 & 52.4 & 45.8 \\
Kangaroo$\dag$~\cite{kangaroogroup} & 7B & 64 & 44.4 & 66.1 & 55.3 & 46.7 & 56.0 & 68.0 & 55.4 & 49.3 & 57.6 & 52.7 \\
LongVA~\cite{longva} & 7B & 128 & 41.1 & 61.1 & 50.4 & 46.2 & 52.6 & 61.6 & 53.6 & 47.6 & 54.3 & 49.3 \\
LongVILA$\dag$~\cite{longvila} & 7B & 256 & 49.0 & 69.3 & 56.1 & 47.0 & 57.5 & 70.4 & 59.2 & 52.1 & 60.6 & 55.7 \\
Video-CCAM~\cite{fei2024videoccam} & 14B & 96 & 42.9 & 62.2 & 50.6 & 46.7 & 53.2 & 66.0 & 56.3 & 49.9 & 57.4 & 51.2 \\
LongLLaVA~\cite{wang2024longllava} & 7B & 256 & - & 61.9 & 51.4 & 45.4 & 52.9 & 66.2 & 54.7 & 50.3 & 57.1 & - \\
Video-XL~\cite{shu2024videoxl} & 7B & 256 & 45.5 & 64.0 &  53.2 &49.2 &55.5 &67.4& 60.7  & 54.9 & 61.0 & 54.0\\

\rowcolor{gray!15}\multicolumn{13}{c}{\textit{RAG-based VLMs}} \\
$\text{Goldfish}\dag$~\cite{ataallah2024goldfish} & 7B & - & 37.3 & 28.7 & 29.4 & 28.7 & 28.9 & 28.6 & 26.4 & 27.3 & 27.4 & 31.2 \\
\text{SALOVA-Qwen}~\cite{kim2024salovasegmentaugmentedlongvideo} & 7B & 1 fps & - & 52.3 & 50.9 & 46.8 & 50.0 & - & - & - & - & -\\

$\text{RAG}_\text{simple}$ & 7B & 128 & 54.9 & \underline{73.7} & 59.2 & 52.0 & 61.6 & 74.6 & 63.2 & 53.6 & 63.8 & 60.1 \\
\midrule

\textbf{$\text{MemVid}$} & 7B & 128 & \textbf{58.1} & \textbf{73.9} & \textbf{63.1} & \textbf{54.1} & \textbf{63.7} & \textbf{75.4} & \textbf{64.6} & \textbf{57.1} & \textbf{65.7} & \textbf{62.5} \\

\bottomrule
\end{tabular}
\end{table*}

\begin{table}[t!]
\centering
\addtolength\tabcolsep{-2.4pt} 
\small
\caption{Experimental results on LVBench. $\dag$ indicates that results are reproduced using their official weights.}
\label{tab:lvbench_simple}
\begin{tabular}{l|c|c|c}
\toprule
\textbf{Model} & \textbf{Size} & \textbf{\# Frame} & \textbf{Overall} \\
\midrule
\rowcolor{gray!15}\multicolumn{4}{c}{\textit{\textcolor{gray}{Proprietary Models} }} \\
\textcolor{gray}{GPT-4o(2024-05-13)~\cite{gpt4o}} & \textcolor{gray}{- }& \textcolor{gray}{10} & \textcolor{gray}{30.8} \\
\textcolor{gray}{Gemini 1.5 Pro~\cite{reid2024gemini}} & \textcolor{gray}{-} & \textcolor{gray}{3600} & \textcolor{gray}{33.1} \\
\textcolor{gray}{GLM-4V-Plus~\cite{glm2024chatglmfamilylargelanguage}} & \textcolor{gray}{-} & \textcolor{gray}{30} & \textcolor{gray}{38.3} \\
\rowcolor{gray!15}\multicolumn{4}{c}{\textit{Open-source VLMs }} \\
MovieChat~\cite{moviechat2023} & 7B & >10000 & 22.5 \\
LLaMA-VID~\cite{llama-vid2023} & 13B & >10800 & 23.9 \\
PLLaVA~\cite{xu2024pllava} & 34B & 16 & 26.1 \\
LLaVA-NeXT-Video~\cite{liu2024llavanextvideo} & 34B & 32 & 32.2 \\
Qwen2-VL$\dag$~\cite{qwen2vl} & 7B & 128 & 37.2 \\
$\text{RAG}_\text{simple}$ & 7B & 128 & 42.0 \\
\midrule
\textbf{MemVid} & 7B & 128 & \textbf{44.4} \\
\bottomrule
\end{tabular}
\vspace{-0.4cm}
\end{table}

\section{Experiments}\label{sec:exp}
\subsection{Experimental Settings}

\subsubsection{Benchmark and Metrics}
We conducted comprehensive experiments on three long video benchmarks characterizing with different features to provide a multi-dimensional evaluation:

\noindent \textbf{VideoMME}~\cite{videomme} consists of 2,700 expert-curated questions linked to 900 diverse videos of different lengths: short (up to 2 minutes), medium (4 to 15 minutes), and long (30 to 60 minutes). It provides two versions of questions, with subtitles and without subtitles. 

\noindent \textbf{MLVU}~\cite{zhou2024mlvu} is a comprehensive video dataset that includes videos ranging from 3 minutes to 2 hours in duration. It encompasses a diverse set of nine tasks, including action recognition, event localization, and counting, designed to evaluate both global and local video understanding. 

\noindent \textbf{LVBench}~\cite{wang2024lvbench} is designed for extremely long videos, with an average duration of 4,101 seconds. It features a diverse set of tasks, including key information retrieval, event understanding, temporal grounding and so on, all supported by high-quality human annotations.

The selection of these three benchmarks provides a comprehensive evaluation framework: VideoMME offers fine-grained analysis across different video lengths, MLVU tests diverse task capabilities, while LVBench focuses on ultra-long video processing. Together they assess model performance from multiple dimensions including video duration, task variety, and understanding depth, ensuring a thorough evaluation of method effectiveness.

\subsubsection{Baselines}
We evaluate MemVid against a wide range of baselines, which are categorized into three groups:

\textbf{1. Proprietary Models}: This category includes state-of-the-art closed-source models such as GPT-4V~\cite{openai2023gpt4}, GPT-4o~\cite{gpt4o}, and Gemini-1.5-Pro~\cite{reid2024gemini}, which have demonstrated strong performance in multi-modal tasks. While these models achieve high scores, their closed nature limits direct architectural comparisons.  

\textbf{2. Open-source VLMs}: We compare against general-purpose Open-source VLMs~\cite{li2023videochat,videollava,chen2024sharegpt4video,qwen2vl,internvl}, which represent the average performance of video understanding models. Additionally, we further include state-of-the-art long-context VLMs~\cite{kangaroogroup,longva,longvila,fei2024videoccam,wang2024longllava,shu2024videoxl}, which extend the context length of traditional VLMs, enabling them to process longer videos effectively.

\textbf{3. RAG-based VLMs}: Relevant works include Goldfish~\cite{ataallah2024goldfish}, SALOVA-Qwen~\cite{kim2024salovasegmentaugmentedlongvideo}, and Video-RAG~\cite{luo2024videorag}. Goldfish converts videos into a text corpus and retrieve key text while SALOVA-Qwen leverages a moment retrieval router to locate key moments. Video-RAG relies on audio and subtitles rather than retrieving key moments, making it orthogonal to our approach, so we do not compare against it. Additionally, we implement a $\text{RAG}_\text{simple}$ variant as a reference baseline, identical to our model but without the memory module.

\subsubsection{Implementation Details}

Our pipeline begins with momentation into 10-second moments using LanguageBind-Large~\cite{zhu2023languagebind} for moment retrieval and subtitle retrieval. The memory model generates four query-aware clues and a draft answer along with the original question, then retrieves the top four moments (reordered chronologically and sampled at 1 fps), as well as the top four subtitle moments. Inputs are truncated to 128 frames with a \(\alpha\) of 0.6, which is aligned with the closest baseline Qwen2VL-7B~\cite{qwen2vl}. For $\text{RAG}_\text{simple}$, we retrieve the top-4 moments according to the question. For Qwen2VL, we sample frames densely from the original video at a rate of 0.25 fps. For supervised fine-tuning, we generate 10,000 synthetic clues and draft answers using Qwen2VL-72B from TVQA-Long~\cite{ataallah2024goldfish}, NExT-QA~\cite{xiao2021nextqa}, and ActivityNet-QA~\cite{yu2019activitynetqa}. We filter out clues that correctly answer the questions and use them as high-quality training labels. For DPO training, we sample different answers from CinePile~\cite{rawal2024cinepile} dataset. We generate diverse clue variations for each sample by adding a little difference in the prompt and evaluate their quality using a frozen VLM's confidence scores on correct answers. We select 1,000 high-quality training pairs where positive clues score above 0.7 and negative clues less than 0.3. Experiments run on a single node with 8×A800 (80GB) GPUs.

\subsection{Overall Results}  
We evaluate MemVid against baseline models on three long-video benchmarks (Table~\ref{tab:main} and Table~\ref{tab:lvbench_simple}). Overall, MemVid establishes new state-of-the-art performance among 7B models, outperforming both conventional long video VLMs and specialized RAG approaches. Specifically:

(1) \textbf{Compared with brute-force down-sampling, MemVid precisely identifies crucial frames for question answering, boosting accuracy under the same context limits.} MemVid leverages the same generation model and input frame limits as Qwen2VL~\cite{qwen2vl} but achieves a +8.7\% absolute gain on MLVU and +6.2\%/+3.7\% improvements on VideoMME (long) without/with subtitles. This validates the effectiveness of MemVid’s memory-enhanced retrieval augmentation framework.

(2) \textbf{MemVid surpasses specialized long-video VLMs with fewer input frames and higher performance.} To enhance VLMs' long-video understanding ability, some models attempt to extend context length to reduce information loss. For example, Video-XL~\cite{shu2024videoxl} and LongVILA~\cite{longvila} can process up to 256–1024 frames. However, MemVid outperforms all long-video VLMs by a significant margin. Specifically, it surpasses Video-XL by +8.2\% and +4.7\% in absolute scores on VideoMME without and with subtitles, respectively. This demonstrates that while current long-video VLMs can process more context, the substantial information loss from compression makes them less effective than expected. In contrast, RAG-based methods like \(\text{RAG}_\text{simple}\) and MemVid show greater potential in tackling long-video understanding tasks.  

\begin{figure}[!t]
    \centering
    \includegraphics[width=0.8\linewidth]{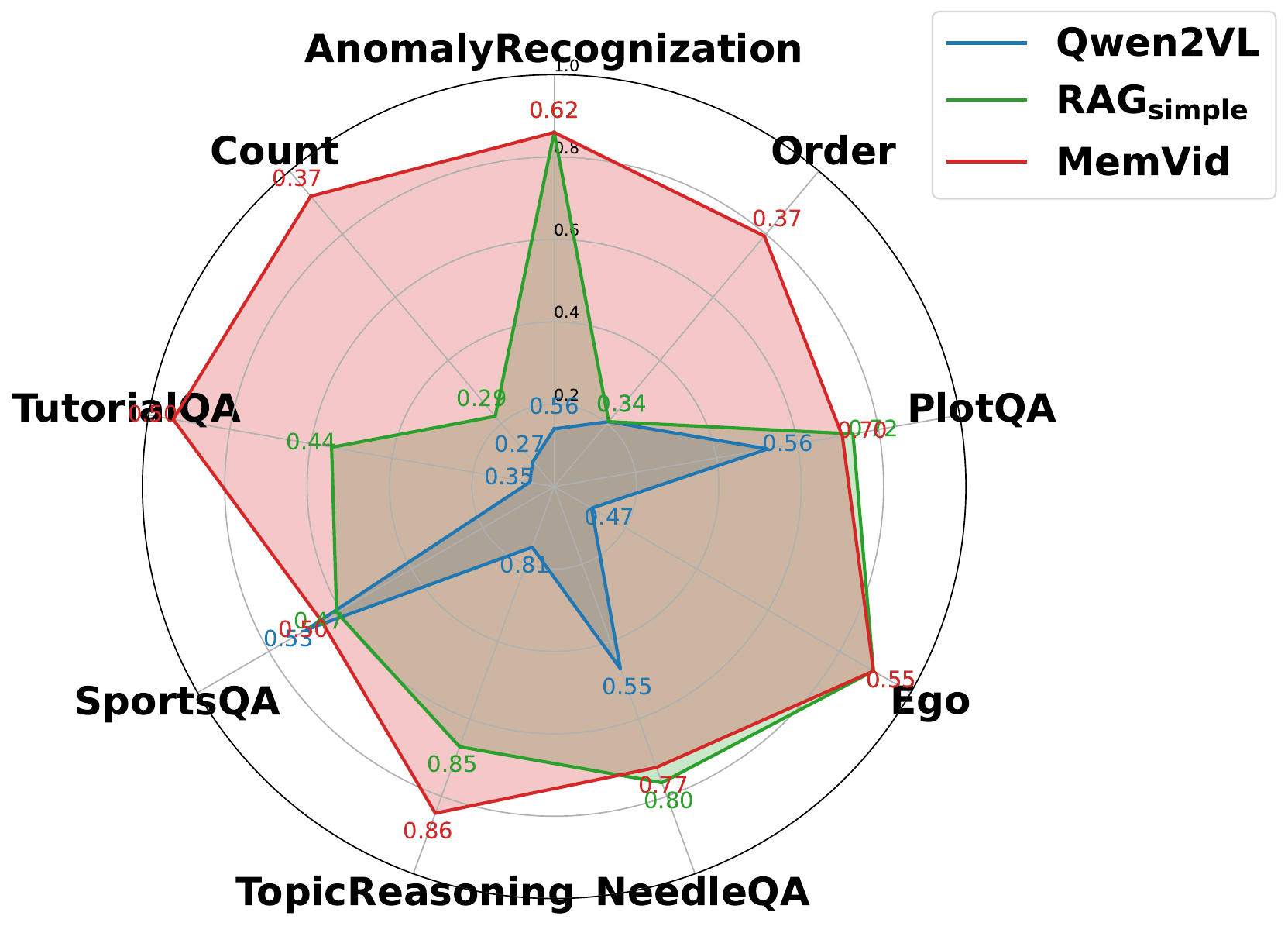}
    \vspace{-0.2cm}
    \caption{Performance comparison of different tasks on MLVU.}
    \label{fig:task}
    \vspace{-0.6cm}
\end{figure}

Additionally, compared to costly proprietary models, MemVid offers a more efficient solution. On VideoMME, although proprietary models like Gemini 1.5 Pro outperform MemVid, they do so at a much higher computational cost (e.g., Gemini 1.5 Pro consumes 928.8k tokens for a 1-hour video, while our model requires only 8.5k tokens). On LVBench, MemVid even surpasses Gemini 1.5 Pro, with 30× less input frames and an absolute gain of 11.3\%, highlighting our model's strong potential for cost-effective long-video understanding.

(3) \textbf{The memory-reasoning mechanism is crucial for long-video understanding.} RAG-based VLMs employ different retrieval strategies and architectures to achieve optimal performance. For instance, SALOVA-Qwen uses a dynamic routing mechanism and spatio-temporal projector to efficiently retrieve and process relevant video segments based on user queries, while Goldfish converts video segments into dense captions for retrieval. Our baseline, \(\text{RAG}_\text{simple}\), uses the same retriever and generation model as MemVid to isolate their impact. However, all these baselines fall far behind our proposed memory-reasoning mechanism. MemVid outperforms the best RAG-based VLM, \(\text{RAG}_\text{simple}\), by +5.8\%/+3.4\%/+3.0\% on MLVU and VideoMME without/with subtitles, underscoring the superiority of our global memory-enhanced clue generation over conventional retrieval augmentation.

Overall, MemVid advances long-video understanding by efficiently identifying key frames through memory-enhanced reasoning, showcasing the strength of memory-augmented retrieval.

\begin{table}[!t]
 \center
 \small
 \caption{Ablation study on MLVU and VideoMME (long).}
  \label{tab:ablation}
     \vspace{-0.3cm}
  \begin{tabular}{lcccc}
  	\toprule
    \multirow{2}{*}{\textbf{Model}}  & \textbf{MLVU} & \multicolumn{2}{c}{\textbf{VideoMME}}  \\
    & \textbf{M-avg} & \textbf{w/o sub.} & \textbf{w sub.}  \\
    \midrule
  	MemVid & \textbf{58.1} & \textbf{54.1} & \textbf{57.1} \\
    \midrule
  	\quad \textit{w/o.} reasoning & 54.9 & 51.6 & 53.6\\
  	\quad \textit{w/o.} memory& 56.2 & 52.4 & 53.8 \\
  	\quad zero-shot& 55.2 & 52.6 & 54.0 \\ 
  	\quad only SFT& 56.4 & 53.3 & 54.9   \\
  	\quad only DPO& 56.6 & 53.7 & 55.6   \\
  	\bottomrule
  \end{tabular}  
  \vspace{-0.4cm}
\end{table}

\subsection{Ablation Study}\label{sec:ablation} 
\begin{sloppypar}
We conduct an ablation study to evaluate MemVid's design by comparing it against several variants. The baseline $\text{MemVid}_\text{w/o reasoning}$ adopts a RAG approach that generates draft answers for retrieval without reasoning about retrieval clues, following methods from language understanding~\cite{gao2022hyde}, while $\text{MemVid}_\text{w/o memory}$ bypasses video scanning entirely and retrieves information using only the original question. To validate our curriculum learning framework, we also examine $\text{MemVid}_\text{zero-shot}$, $\text{MemVid}_\text{onlySFT}$ (supervised fine-tuning only), and $\text{MemVid}_\text{onlyDPO}$ (DPO fine-tuning only). The results in Table~\ref{tab:ablation} demonstrate that MemVid's memory mechanism provides significant advantages: (1) MemVid achieves superior zero-shot performance compared to both $\text{MemVid}_\text{w/o reasoning}$ and $\text{MemVid}_\text{w/o memory}$, with the latter showing intermediate improvements of 2.4\% on MLVU and 1.6\%/0.4\% on VideoMME (without/with subtitles). This verify the effectiveness of our framework.  Further performance gains of around 3\% are achieved through the two-stage curriculum learning, confirming the effectiveness of the memory-augmented approach. (2) Our curriculum learning framework also proves impactful, where SFT provides around 1\% improvement over the zero-shot baseline, and subsequent DPO training yields an additional 2\% gain by refining the effectiveness of retrieval clues for downstream long-video understanding.
\end{sloppypar}

\subsection{Generalizability Analysis}
To thoroughly assess MemVid's generalization capability, we conduct extensive experiments across diverse task scenarios, varying context constraints, and diverse downstream model architectures.

\subsubsection{Task-Specific Performance}  
To analyze the performance of MemVid across different tasks, we compare it with $\text{Qwen2VL}$ and $\text{RAG}_\text{simple}$ in Figure~\ref{fig:task}. The results demonstrate that whether on tasks with explicit queries (NeedleQA, Count, and Order) or complex tasks without explicit queries (TutorialQA, TopicReasoning, SportsQA), and whether on single-detail question answering (NeedleQA) or multiple-detail question answering (Order, Count), MemVid robustly shows a significant advantage over $\text{RAG}_\text{simple}$. Specifically, MemVid outperforms $\text{RAG}_\text{simple}$ by an absolute score of 8\% on the Count task, 6\% on TutorialQA, and 3\% on the Order task. For many tasks, such as Order and TutorialQA, our training set does not include similar training data. Nevertheless, the model generalizes well and exhibits strong performance on these tasks, which demonstrates the effectiveness of our proposed memory architecture.

\begin{figure}[!t]
    \centering
    \includegraphics[width=0.8\linewidth]{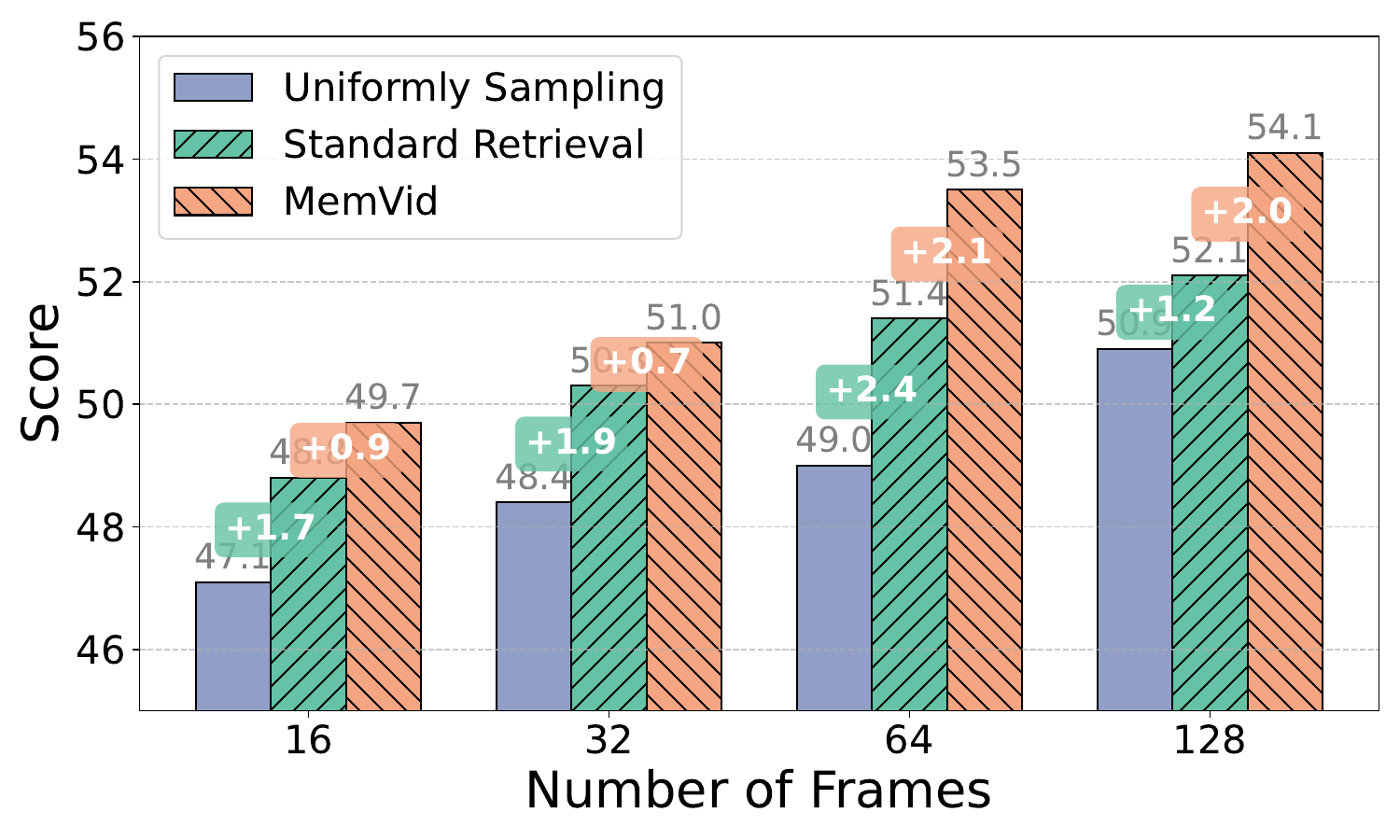}
  \vspace{-0.4cm}
    \caption{Performance of different frame numbers of downstream VLM, evaluated on VideoMME (long).}
    \label{fig:frame}
  \vspace{-0.3cm}
\end{figure}

\subsubsection{Frame Number Analysis}  
As shown in Figure~\ref{fig:frame}, we evaluate three frame-selection strategies: uniformly sampling, standard retrieval, and MemVid. These strategies leverage the same generative model, Qwen2VL-7B, for video question answering, with inputs of 16, 32, 64, and 128 frames. As shown in Figure~\ref{fig:frame}, \textbf{MemVid consistently outperforms uniformly sampling strategy and standard retrieval strategy}, with performance gains of 0.9, 0.7, 2.1, and 2.9 percentage points at 16, 32, 64, and 128 frames, respectively. Additionally, we can find that MemVid's advantage grows as frame number increases. This is because when fewer than 32 frames are provided, the retrieved clips must be aggressively downsampled, limiting MemVid’s effectiveness. In contrast, with 64 or more frames, the model leverages its full retrieval capability, leading to more significant performance improvements.  

\begin{figure*}[!t]
    \centering
    \includegraphics[width=0.9\textwidth]{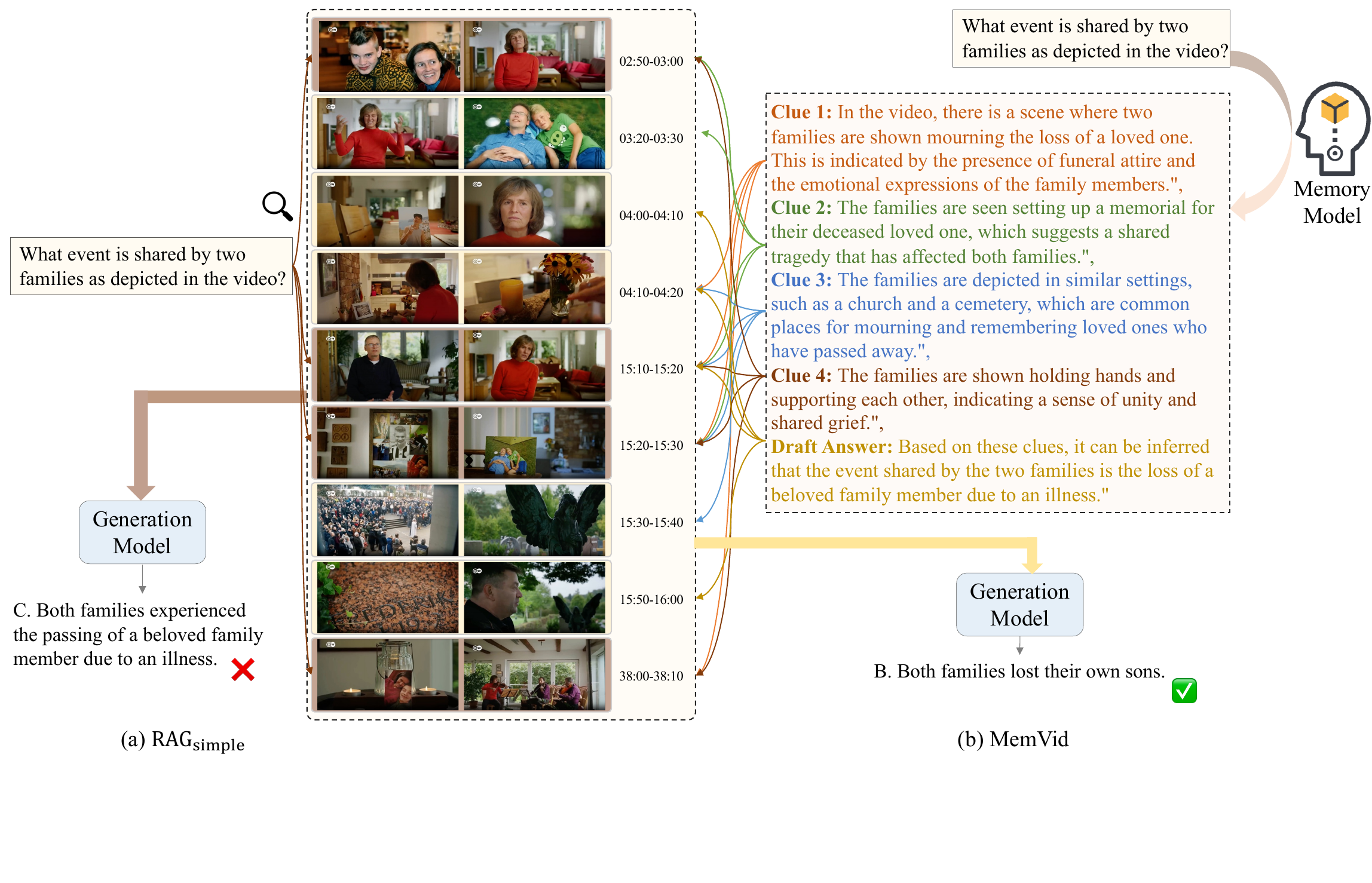}
  \vspace{-0.3cm}
    \caption{Visualization of $\text{RAG}_\text{simple}$ and MemVid on VideoMME, with arrows highlighting the retrieved video clips for each query.}
    \label{fig:visual}
  \vspace{-0.4cm}
\end{figure*}

\begin{table}[!t]
  \centering
  \small
  \caption{Zero-shot application to different downstream generation models with different sizes.}
  \label{tab:downstream}
  \vspace{-0.3cm}
  \begin{tabular}{lccccc}
    \toprule
    \textbf{Model} & \textbf{Size} & \textbf{\# Frame}  & \textbf{Performance} & \textbf{Gain}  \\
    \midrule
    VILA-1.5 & 3B & 8& 32.7 & - \\
     + MemVid & 3B &8 &  36.0  & +10.1\% \\
    \hline
    LongVA & 7B & 128 & 42.4 & - \\
    + MemVid& 7B & 128&  43.9 & +3.5\% \\
    \hline
    Qwen2VL & 72B  & 128 &  59.0 & - \\
    + MemVid & 72B  & 128 &  60.4 & +2.4 \%\\
    \bottomrule
  \end{tabular}  
  \vspace{-0.4cm}
\end{table}

\subsubsection{Different Downstream Architectures} 
Our memorizer and retriever are modular components compatible with diverse downstream models. We evaluate their effectiveness across VLMs of varying architectures and scales,
including VILA1.5-3B~\cite{lin2024vila}, LongVA-7B-DPO~\cite{longva}, and Qwen2VL-72B~\cite{qwen2vl}. As shown in Table~\ref{tab:downstream}, MemVid improves performance across all models, achieving a 10.1\% gain on VILA-1.5 (3B, 8 frames), 3.5\% on LongVA (7B, 128 frames), and 2.4\% on Qwen2VL (72B, 128 frames). The largest gains on smaller models under limited inputs suggest that MemVid effectively locates the most important frames, compensating for weaker architectures. Notably, these gains generalize across models despite MemVid being trained solely with Qwen2VL-7B feedback, demonstrating its adaptability without model-specific tuning.

In summary, MemVid demonstrates strong generalization across diverse tasks (explicit and implicit queries, single-detail and multi-detail question-answering samples), model architectures (3B-72B VLMs), and frame constraints (16-128), consistently outperforming baselines. This highlights the value of our proposed memory-augmented retrieval mechanism.

\begin{table}[!t]
  \centering
  \small
  \caption{Efficiency comparison of long VLM and MemVid on VideoMME(long) w/o subtitle.}
  \label{tab:efficiency}
  \vspace{-0.2cm}
  \begin{tabular}{lccc}
    \toprule
    \textbf{Metric} & \textbf{VideoXL} & \textbf{MemVid} & \textbf{Gain} \\
    \midrule
    \textbf{\# Frame} & 1024 & 64 & \textbf{$\downarrow$ 93.8\%}\\
    \textbf{Latency(s) } & 85.0 & 55.2 &\textbf{$\downarrow$ 35.1\%}\\
    \textbf{Memory (GB) } & 56.1 & 36.5 & \textbf{$\downarrow$ 34.9\%}\\
    \textbf{Performance } & 45.6 & 53.5 &  \textbf{$\uparrow$ 17.3\%}\\
    \bottomrule
  \end{tabular}  
  \vspace{-0.4cm}
\end{table}

\subsection{Efficiency}
From Table~\ref{tab:main}, we can conclude that MemVid surpasses specialized long-video VLMs by achieving higher performance with fewer input frames. To evaluate MemVid's efficiency in detail, we compare it with a top-performing LVLM, VideoXL, across multiple dimensions, including the number of input frames, latency, memory usage, and performance. As shown in Table~\ref{tab:efficiency}, MemVid significantly reduces input length by selecting the most informative frames, allowing downstream model to process only 64 frames while reducing GPU memory usage by 34.9\% and inference latency by 35.1\%, yet achieving 17.3\% higher performance compared to VideoXL with 1024 frames. These results highlight that increasing the frame number introduces higher computational cost but still suffers from information loss, whereas MemVid efficiently extracts a compact yet informative subset, enabling faster inference and superior performance.

 \subsection{Case Study Analysis}  
Figure~\ref{fig:visual} shows a case demonstration of different methods for answering the question, ``What event is shared by two families?'' The baseline model $\text{RAG}_\text{simple}$ retrieves moments directly but fails to capture critical evidence, such as the father and son lying together (indicating their relationship) and the church mourning scene (revealing the families' shared loss). In contrast, MemVid leverages context-aware reasoning to infer implicit semantics and decomposes the problem into fine-grained and explicit clues, which helps retrieve comprehensive evidence and facilitates the generation model in inferring the precise answer.

\section{Conclusion}

In this paper, we propose MemVid, a novel RAG-based LVU approach inspired by human cognitive memory. MemVid operates through four key steps: memorizing holistic video information, reasoning about task-specific information needs, retrieving critical moments, and focusing on them to generate the final answer. To improve memory-grounded reasoning and optimize performance, we introduce a curriculum learning strategy. Extensive experiments on LVU benchmarks (MLVU, VideoMME, LVBench) demonstrate that MemVid significantly outperforms existing RAG-based methods and popular LVU models, validating its effectiveness.

\newpage
{\small
\bibliographystyle{unsrt}
\bibliography{egbib}
}

\newpage

\end{document}